\documentclass[11pt,a4paper]{article}
\usepackage[utf8]{inputenc}
\usepackage{amsmath}
\usepackage{amsfonts}
\usepackage{amssymb}
\usepackage{graphicx}
\usepackage[hypertexnames=false]{hyperref}
\usepackage{cite}
\usepackage{float}
\usepackage{tikz}
\usetikzlibrary{shapes,arrows,positioning,decorations.pathreplacing}

\title{Phionyx: A Deterministic AI Runtime Architecture with Structured State Management and Pre-Response Governance}

\author{Ali Toygar Abak\\Phionyx Research\\{\small\texttt{founder@phionyx.ai}}}

\date{April 2026}

\begin{document}

\maketitle

\begin{abstract}
We present Phionyx, a deterministic AI runtime architecture derived from the broader Echoism interaction framework that introduces a governance-first approach to AI engineering: treating large language model (LLM) outputs as \textbf{noisy sensor measurements} rather than direct decisions. Unlike probabilistic agents, Phionyx enforces deterministic state evolution via a structured state vector governed by deterministic state-evolution equations, enabling reproducible behavior in applications requiring auditability and governance. The architecture integrates three layers: (1) a deterministic evaluation kernel processing noisy sensor measurements through a canonical 46-block pipeline, (2) a unified safety layer providing pre-response control and architectural privacy enforcement, and (3) a semantic time-based memory system implementing impact-weighted cache eviction. Experimental validation on single-instance deployments demonstrates approximately 31\% reduction in computational overhead vs.\ post-hoc filtering (at 30\% unsafe input ratio, simulated cost model) and up to 24\% improvement in high-value data retention vs.\ LRU (72\% vs.\ FIFO, same cache capacity, benchmark-verified), deterministic execution verified across 100 repeated runs with zero variance in control signals (hash-verified), and zero unplanned restarts in single-instance deployment testing (see Appendix~C for methodology and scope). This paper presents the architecture, its analytic structure, and scoped experimental evidence; generalization to distributed or multi-tenant deployments remains future work.
\end{abstract}

\section{Introduction}

\subsection{Motivation}

Current artificial intelligence systems, particularly those based on large language models (LLMs), suffer from fundamental limitations that prevent their deployment in regulated, auditable, or safety-sensitive applications:

\begin{enumerate}
\item \textbf{Non-determinism}: The same input may produce different outputs on different executions, making these systems unsuitable for applications requiring reproducibility and auditability.
\item \textbf{Static State Representation}: System state is typically represented as static data structures (JSON objects, key-value stores, vector embeddings) that are stored and retrieved but do not evolve according to governing equations, failing to model dynamic state transitions required for behavioral governance.
\item \textbf{Post-hoc Safety Mechanisms}: Governance mechanisms operate as post-hoc filters or output censors, applying safety checks after responses are generated, failing to prevent risky behavior at its source.
\item \textbf{Chronological Time Models}: Memory systems use chronological timestamps or simple decay functions that do not account for the semantic or cognitive significance of information.
\item \textbf{Privacy and Isolation Challenges}: Multi-participant systems share state or context, leading to cross-participant state leakage where one participant's cognitive state influences another's.
\end{enumerate}

Existing approaches---including guardrails frameworks~\cite{nemoguardrails,guardrailsai}, constrained-generation languages~\cite{lmql,guidance}, and agent orchestrators~\cite{langchain,autogen}---address subsets of these problems but do not provide a unified runtime layer that combines deterministic state evolution, pre-response governance, structured state management, and participant isolation in a single architecture. This paper fills that gap.

\subsection{Contributions}

This paper presents Phionyx, a deterministic AI runtime architecture. Its design draws on the Echoism ontological framework, which models interaction through resonance, feedback, and transformation; Phionyx operationalizes a subset of these principles as a concrete engineering runtime. The architecture addresses the limitations above through three integrated innovations:

\begin{enumerate}
\item \textbf{Deterministic Evaluation Kernel}: A canonical pipeline of evaluation blocks that treats LLM outputs as noisy sensor measurements and generates deterministic control signals based on structured state metrics.
\item \textbf{Unified Safety \& Governance Layer}: An architectural security layer providing pre-response internal state governance, cognitive envelopes for secure agent communication, non-persistence doctrines for derived metrics, and participant-scoped cognitive isolation.
\item \textbf{Semantic Time-Based Memory System}: An impact-weighted cache eviction strategy that models time as a semantic vector affecting cognitive impact, improving high-value data retention by up to 24\% vs.\ LRU and up to 72\% vs.\ FIFO (benchmark-verified, same cache capacity).
\end{enumerate}

This paper is a \textbf{systems architecture paper}: it describes the design, analytic structure, and scoped experimental validation of a concrete runtime system. It is not a cognitive architecture proposal, a benchmark study, or a product whitepaper.

\section{Related Work}

\subsection{Deterministic AI Systems}

Previous work on deterministic AI has focused on rule-based systems~\cite{rulebased}, symbolic reasoning~\cite{symbolic}, and constraint satisfaction~\cite{constraint}. However, these approaches lack the creative capabilities of probabilistic models. Work on neural network calibration~\cite{calibration} and LLM self-knowledge~\cite{llmknowledge} establishes that model outputs carry quantifiable uncertainty, motivating our treatment of LLM outputs as noisy sensor measurements. However, prior work does not provide deterministic decision-making or structured state management over these uncertain outputs.

\subsection{AI Safety and Governance}

Safety mechanisms in AI systems typically operate post-hoc~\cite{posthoc1,posthoc2}, applying filters or censors after response generation. Constitutional approaches~\cite{constitutional} and input-output safeguards~\cite{llamaguard} move safety closer to the generation process but do not integrate with structured state or provide architectural enforcement. Our work extends this by providing deterministic pre-response control based on operational constraint vectors computed from structured state.

\subsection{Memory and Time Models}

Memory systems in AI typically use recency-weighted retrieval~\cite{generativeagents}, attention-based selection~\cite{memorynetworks}, or simple exponential decay~\cite{exponential}. Differentiable memory architectures~\cite{ntm,endtoendmem} have been proposed but do not integrate with structured state or provide impact-weighted cache eviction. Our semantic time model provides deterministic decay affecting state metrics (entropy, Phi) and enabling autonomous cache eviction.

\subsection{Privacy and Isolation}

Multi-agent systems typically share state or context~\cite{multistate1,multistate2}, leading to cross-participant state leakage. Data isolation through federated approaches~\cite{federated,fedsurvey} addresses training-time privacy but does not provide participant-scoped state isolation at runtime with architectural enforcement. Our architecture maintains separate state vectors per participant in isolated containers.

\subsection{Structured LLM Orchestration and Guardrails}

Recent frameworks address LLM control through complementary mechanisms. NeMo Guardrails~\cite{nemoguardrails} provides programmable input/output rails via a domain-specific language (Colang), operating at the prompt and response boundary. LMQL~\cite{lmql} and Guidance~\cite{guidance} enable constrained decoding and structured generation within the LLM call itself. DSPy~\cite{dspy} compiles declarative language model calls into optimized prompt pipelines. Guardrails AI~\cite{guardrailsai} validates LLM outputs against user-defined schemas. Agent orchestrators such as LangChain~\cite{langchain} and AutoGen~\cite{autogen} coordinate multi-step LLM workflows with tool access and memory.

These frameworks address prompt engineering, output validation, and constrained generation, but none provides (a)~deterministic state evolution governed by analytic equations, (b)~pre-response governance that modulates control signals \textit{before} the LLM is invoked for unsafe inputs, (c)~structured state vectors whose metrics (entropy, valence, amplitude) actively drive safety gates and cache eviction, or (d)~per-participant state isolation enforced at the architectural level. Phionyx operates at a different layer: between the LLM and the application, governing the entire decision path with deterministic state-evolution logic rather than constraining individual LLM calls.

\subsection{Positioning Summary}

Table~\ref{tab:comparison} summarizes how Phionyx differs from adjacent system categories along four dimensions: determinism, pre-response governance, structured state, and participant isolation.

\begin{table}[ht]
\centering
\small
\begin{tabular}{|l|c|c|c|c|}
\hline
\textbf{System Category} & \textbf{Determ.} & \textbf{Pre-resp.} & \textbf{Struct.} & \textbf{Isol.} \\
 & \textbf{Control} & \textbf{Gov.} & \textbf{State} & \\
\hline
Post-hoc filters~\cite{posthoc1,llamaguard} & -- & -- & -- & -- \\
Constitutional AI~\cite{constitutional} & -- & partial & -- & -- \\
Agent orchestrators~\cite{langchain,autogen} & -- & -- & partial & -- \\
Tracing / observability~\cite{langsmith,phoenix} & -- & -- & -- & -- \\
Policy engines / guardrails~\cite{nemoguardrails,guardrailsai} & -- & partial & -- & -- \\
Federated learning~\cite{federated} & -- & -- & -- & partial \\
\textbf{Phionyx (this work)} & \checkmark & \checkmark & \checkmark & \checkmark \\
\hline
\end{tabular}
\caption{Comparison of Phionyx with adjacent system categories. \textbf{Determ.\ Control}: deterministic, reproducible control signals from identical inputs. \textbf{Pre-resp.\ Gov.}: governance decisions made before response generation. \textbf{Struct.\ State}: state evolves via governing equations, not static storage. \textbf{Isol.}: per-participant state isolation enforced architecturally.}
\label{tab:comparison}
\end{table}

The key distinction is that tracing tools, post-hoc filters, and policy engines operate \textit{after or alongside} the generation process, while Phionyx's pipeline sits \textit{between} the LLM and the response, governing the entire path from input to output with deterministic state-evolution logic. This is not merely ``deterministic logging''---the state vector actively modulates response amplitude, triggers safety gates, and drives cache eviction decisions before any output is delivered.

\section{Architecture Overview}

Phionyx consists of three integrated layers operating on a unified structured state vector.

\begin{figure}[h]
\centering
\begin{tikzpicture}[node distance=2cm, auto, scale=0.8, every node/.style={transform shape}]
    \node [rectangle, draw, fill=blue!20, minimum width=3cm, minimum height=1.5cm, align=center] (layer1) {Layer 1: Deterministic\\Cognitive Kernel\\(46 Blocks)};

    \node [rectangle, draw, fill=red!20, minimum width=3cm, minimum height=1.5cm, align=center, below of=layer1, yshift=-1cm] (layer2) {Layer 2: Safety \&\\Governance Layer\\(Pre-response, Envelopes)};

    \node [rectangle, draw, fill=green!20, minimum width=3cm, minimum height=1.5cm, align=center, below of=layer2, yshift=-1cm] (layer3) {Layer 3: Semantic Time\\Memory System\\(Cache Eviction)};

    \node [ellipse, draw, fill=yellow!20, minimum width=2cm, minimum height=1cm, align=center, right of=layer2, xshift=3cm] (state) {State Vector\\$\mathbf{S}(t)$};
    
    \draw [<->, thick] (layer1) -- (state);
    \draw [<->, thick] (layer2) -- (state);
    \draw [<->, thick] (layer3) -- (state);
    \draw [->, thick] (layer1) -- (layer2);
    \draw [->, thick] (layer2) -- (layer3);
\end{tikzpicture}
\caption{Three-layer Phionyx architecture: (1) Deterministic Cognitive Kernel with canonical pipeline, (2) Safety \& Governance Layer with pre-response control and cognitive envelopes, (3) Semantic Time-Based Memory System with impact-weighted cache eviction. All layers interact with the unified structured state vector $\mathbf{S}(t)$.}
\label{fig:architecture}
\end{figure}
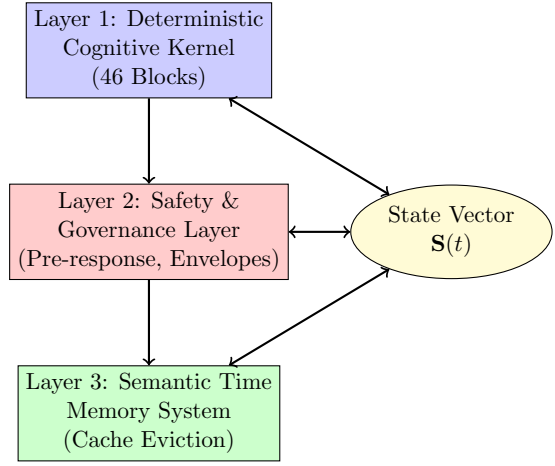

\subsection{System State Representation}

The system maintains a two-tier state representation distinguishing \textit{primary} (persistent) state from \textit{derived} (computed-on-demand) metrics.

\textbf{Primary state} (persistent, updated each turn):
\begin{equation}
\mathbf{S}_p(t) = [H(t),\; A(t),\; V(t),\; \dot{A}(t),\; \dot{V}(t),\; t_{\text{local}},\; t_{\text{global}}]
\end{equation}

where $H$ is entropy (behavioral stability), $A$ is amplitude (arousal), $V$ is valence (affective polarity, range $[-1, 1]$), $\dot{A}$/$\dot{V}$ are their rates of change, and $t_{\text{local}}$/$t_{\text{global}}$ are semantic time components (Section~6). Additionally, discrete event tags $E_{\text{tags}}$ track categorical state markers.

\textbf{Derived metrics} (non-persistent, computed on-demand):
\begin{equation}
\Phi(t) = f_\Phi(\mathbf{S}_p(t)), \quad R(t) = f_R(\mathbf{S}_p(t))
\end{equation}

where $\Phi$ (cognitive resonance) and $R$ (resonance score) are computed from primary state and \textbf{never stored} (Non-Persistence Doctrine, Section~5.3).

For mathematical exposition in subsequent sections, we use the compact notation $\mathbf{S}(t)$ to denote the full observable state (primary components plus derived metrics). The reference implementation extends the primary vector to include additional operational fields; the paper presents the minimal set sufficient to describe the architecture's analytic structure.

The state vector evolves according to deterministic state-evolution equations (Section~\ref{sec:physics}) rather than simple data storage.

\subsection{Three-Layer Architecture}

\textbf{Layer 1: Deterministic Cognitive Kernel}
\begin{itemize}
\item Canonical pipeline of 46 evaluation blocks
\item Treats LLM outputs as noisy sensor measurements
\item Generates deterministic control signals based on state metrics
\item Implements failure classification and deterministic recovery
\end{itemize}

\textbf{Layer 2: Safety \& Governance Layer}
\begin{itemize}
\item Pre-response internal state governance with operational constraint vectors
\item Cognitive envelopes for secure agent communication
\item Non-persistence doctrines for derived metrics
\item Participant-scoped cognitive isolation
\item Output distortion prevention: CEP engine with 4 detection mechanisms (identity confabulation, distress-language mimicry, repetitive echo loops, self-diagnostic assertions)
\end{itemize}

\textbf{Layer 3: Semantic Time-Based Memory System}
\begin{itemize}
\item Semantic time vector: $\mathbf{T} = [dt, t_{\text{local}}, t_{\text{global}}]$
\item Deterministic decay affecting entropy and derived metrics
\item Cognitive impact degradation with impact weights
\item Monotonic clock ensuring time irreversibility
\end{itemize}

\section{Layer 1: Deterministic Cognitive Kernel}

\begin{figure}[h]
\centering
\begin{tikzpicture}[
    node distance=0.3cm,
    group/.style={rectangle, draw, rounded corners, minimum width=4.8cm, minimum height=0.9cm, font=\small},
    ann/.style={font=\scriptsize\itshape, text=gray},
    scale=0.85, every node/.style={transform shape}
]
    \node [group, fill=red!15]   (G1) {G1: Ingress \& Safety (B1--B5)};
    \node [ann, right=0.2cm of G1] (a1) {kill switch, input gate, intent, RAG};

    \node [group, fill=blue!15, below=of G1]  (G2) {G2: Perception \& State Init (B6--B12)};
    \node [ann, right=0.2cm of G2] (a2) {frames, state init, goals, UKF predict};

    \node [group, fill=orange!15, below=of G2] (G3) {G3: Cognition \& Governance (B13--B22)};
    \node [ann, right=0.2cm of G3] (a3) {self-model, ethics, CEP, narrative};

    \node [group, fill=yellow!15, below=of G3] (G4) {G4: State Evolution \& Causality (B23--B36)};
    \node [ann, right=0.2cm of G4] (a4) {drift, physics, causal graph, world snapshot};

    \node [group, fill=green!15, below=of G4] (G5) {G5: Fusion \& Response (B37--B42)};
    \node [ann, right=0.2cm of G5] (a5) {$\Phi$/entropy compute, confidence, response build};

    \node [group, fill=purple!15, below=of G5] (G6) {G6: Consolidation \& Audit (B43--B46)};
    \node [ann, right=0.2cm of G6] (a6) {memory consolidation, audit, feedback, learning};

    \draw [->, thick] (G1) -- (G2);
    \draw [->, thick] (G2) -- (G3);
    \draw [->, thick] (G3) -- (G4);
    \draw [->, thick] (G4) -- (G5);
    \draw [->, thick] (G5) -- (G6);

    \node [ellipse, draw, fill=yellow!20, minimum width=1.6cm, font=\small,
           left=1.2cm of G3] (sv) {$\mathbf{S}(t)$};
    \draw [<->, densely dashed] (sv) -- (G2);
    \draw [<->, densely dashed] (sv) -- (G3);
    \draw [<->, densely dashed] (sv) -- (G4);
    \draw [<->, densely dashed] (sv) -- (G5);
\end{tikzpicture}
\caption{Canonical pipeline (46 blocks, v3.8.0) organized into six functional macro-groups. Arrows denote deterministic execution order; dashed lines indicate state vector reads and writes. All blocks within a group execute sequentially; no block may be reordered across group boundaries.}
\label{fig:pipeline}
\end{figure}
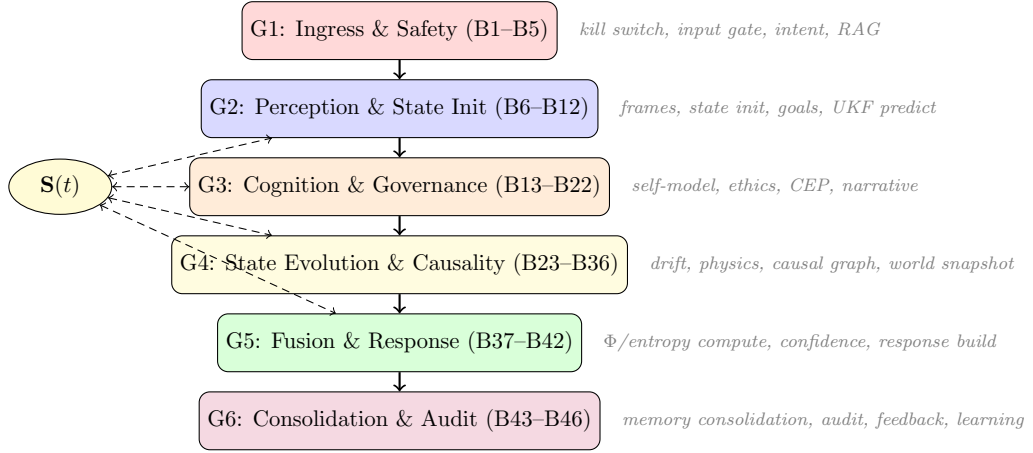

\subsection{Canonical Pipeline}

The cognitive kernel executes a predefined sequence of 46 evaluation blocks in deterministic order:

\begin{equation}
\text{Pipeline} = [B_1, B_2, \ldots, B_{46}]
\end{equation}

where each block $B_i$ performs a specific cognitive evaluation function:

\begin{equation}
B_i: (\mathbf{S}(t), \mathbf{M}(t), \mathbf{I}) \rightarrow (\mathbf{S}(t+\Delta t), \mathbf{C}_i)
\end{equation}

where:
\begin{itemize}
\item $\mathbf{S}(t)$ is the current structured state
\item $\mathbf{M}(t)$ are noisy sensor measurements (LLM outputs)
\item $\mathbf{I}$ is the input
\item $\mathbf{C}_i$ is the control signal generated by block $i$
\end{itemize}

The 46 blocks are organized into six functional macro-groups that reflect the cognitive processing stages (see Figure~\ref{fig:pipeline} and Appendix~A for the full canonical listing):

\begin{enumerate}
\item \textbf{Ingress \& Safety} (B1--B5): Kill switch gate, time update, input safety, intent classification, context retrieval.
\item \textbf{Perception \& State Initialization} (B6--B12): Perceptual framing, scenario creation, state initialization, goal evaluation/decomposition, UKF prediction, pre-response entropy gate.
\item \textbf{Cognition \& Governance} (B13--B22): Cognitive processing, self-model assessment, knowledge boundary, trust evaluation, ethics (pre/post and deliberative), CEP output distortion prevention, narrative generation, action intent gate.
\item \textbf{State Evolution \& Causality} (B23--B36): Behavioral drift detection, state update (physics and ESC), Phi publication, post-response entropy gate, memory growth, emotion estimation, causal graph operations, world state snapshot.
\item \textbf{Fusion \& Response} (B37--B42): Phi/entropy computation, confidence fusion, arbitration resolution, response revision gate, response construction.
\item \textbf{Consolidation \& Audit} (B43--B46): Memory consolidation, audit logging (hash chain + Ed25519), outcome feedback, learning gate.
\end{enumerate}

\subsection{LLM as Noisy Sensor}

A fundamental innovation of Phionyx is treating LLM outputs as \textbf{noisy sensor measurements} rather than direct decisions. This approach enables deterministic AI while preserving the creative capabilities of probabilistic models.

The system models LLM outputs as:

\begin{equation}
\mathbf{M}(t) = f_{\text{LLM}}(\mathbf{I}, \mathbf{S}(t)) + \mathbf{N}(t)
\end{equation}

where $\mathbf{N}(t)$ represents measurement noise (probabilistic variation inherent in LLM outputs). This noise is not a bug but a feature: it captures the creative variability of language models while allowing deterministic processing downstream.

The cognitive kernel then processes these noisy measurements deterministically:

\begin{equation}
\mathbf{C}(t) = g_{\text{deterministic}}(\mathbf{M}(t), \mathbf{S}(t))
\end{equation}

This approach achieves deterministic, reproducible behavior while maintaining the creative capabilities of probabilistic models. The key insight is that \textit{decision-making} should be deterministic, while \textit{creativity} (captured in the noisy sensor measurements) can remain probabilistic. This separation of concerns is fundamental to Phionyx's architecture and represents a distinct approach to AI engineering.

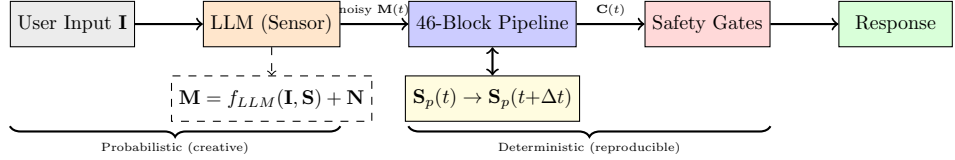
\begin{figure}[ht]
\centering
\begin{tikzpicture}[node distance=1.2cm, auto, scale=0.75, every node/.style={transform shape, font=\small}]
    \node [rectangle, draw, fill=gray!15, minimum width=2cm, minimum height=0.8cm] (user) {User Input $\mathbf{I}$};
    \node [rectangle, draw, fill=orange!20, minimum width=2.2cm, minimum height=0.8cm, right=1.2cm of user] (llm) {LLM (Sensor)};
    \node [rectangle, draw, dashed, minimum width=2.2cm, minimum height=0.8cm, below=0.5cm of llm] (noise) {$\mathbf{M} = f_{LLM}(\mathbf{I},\mathbf{S}) + \mathbf{N}$};
    \node [rectangle, draw, fill=blue!20, minimum width=2.5cm, minimum height=0.8cm, right=1.2cm of llm] (pipeline) {46-Block Pipeline};
    \node [rectangle, draw, fill=yellow!15, minimum width=2cm, minimum height=0.8cm, below=0.5cm of pipeline] (state) {$\mathbf{S}_p(t) \to \mathbf{S}_p(t{+}\Delta t)$};
    \node [rectangle, draw, fill=red!15, minimum width=2.2cm, minimum height=0.8cm, right=1.2cm of pipeline] (gates) {Safety Gates};
    \node [rectangle, draw, fill=green!15, minimum width=2cm, minimum height=0.8cm, right=1.2cm of gates] (resp) {Response};

    \draw [->, thick] (user) -- (llm);
    \draw [->, thick] (llm) -- node[above, font=\tiny]{noisy $\mathbf{M}(t)$} (pipeline);
    \draw [->, thick] (pipeline) -- node[above, font=\tiny]{$\mathbf{C}(t)$} (gates);
    \draw [->, thick] (gates) -- (resp);
    \draw [<->, thick] (pipeline) -- (state);
    \draw [->, dashed] (llm) -- (noise);

    \draw [decorate, decoration={brace, amplitude=4pt, mirror}, thick] ([yshift=-1.8cm]pipeline.west) -- ([yshift=-1.8cm]gates.east) node[midway, below=4pt, font=\tiny]{Deterministic (reproducible)};
    \draw [decorate, decoration={brace, amplitude=4pt, mirror}, thick] ([yshift=-1.8cm]user.west) -- ([yshift=-1.8cm]llm.east) node[midway, below=4pt, font=\tiny]{Probabilistic (creative)};
\end{tikzpicture}
\caption{The noisy sensor separation: LLM outputs ($\mathbf{M}$) are treated as probabilistic measurements. The 46-block pipeline, state evolution, and safety gates operate deterministically on these measurements, producing reproducible control signals $\mathbf{C}(t)$ regardless of LLM variance.}
\label{fig:sensor}
\end{figure}

\subsection{Structured State Evolution}
\label{sec:physics}

The state vector evolves according to deterministic state-evolution equations. These are physics-inspired control equations---engineered update rules with analytic properties (bounded, monotonic decay, deterministic)---not empirical laws of nature:

\textbf{Entropy Evolution:}
\begin{equation}
S(t+\Delta t) = S(t) + \Delta S_{\text{increase}} - \Delta S_{\text{decay}}
\end{equation}

where:
\begin{align}
\Delta S_{\text{increase}} &= \alpha \cdot \beta \cdot dt \\
\Delta S_{\text{decay}} &= S(t) \cdot \gamma \cdot dt
\end{align}

with $\alpha$ = uncertainty factor, $\beta$ = decay rate, $\gamma$ = natural decay rate.

\textbf{Phi Evolution:}
\begin{equation}
\Phi(t+\Delta t) = \Phi(t) \cdot e^{-\lambda \cdot dt}
\end{equation}

where $\lambda$ is the decay coefficient.

\textbf{Amplitude Control:}
\begin{equation}
A(t+\Delta t) = A(t) \cdot \text{gate}(\mathbf{C}(t), \mathbf{S}(t))
\end{equation}

where $\text{gate}()$ is a deterministic gating function based on control signals and state.

\subsection{Failure Classification and Recovery}

Failures are classified based on cognitive state metrics:

\begin{equation}
\text{FailureType} = \text{classify}(S(t), \Phi(t), \mathbf{C}(t))
\end{equation}

Recovery strategies are deterministic and restore the structured state vector:

\begin{equation}
\mathbf{S}_{\text{recovered}} = \text{recover}(\mathbf{S}(t), \text{FailureType})
\end{equation}

This approach reduces computational resource consumption compared to generic retries or system restarts.

\section{Layer 2: Safety \& Governance Layer}

\subsection{Pre-Response Internal State Governance}

Before generating a response, the system measures internal cognitive state and calculates an operational constraint vector:

\begin{equation}
\mathbf{V}_{\text{constraint}}(t) = \text{compute}(S(t), \Phi(t), A(t), \mathbf{C}(t))
\end{equation}

The constraint vector quantifies potential violations of operational policies. When constraints are violated, response amplitude is reduced deterministically:

\begin{equation}
A_{\text{response}}(t) = A(t) \cdot \text{damp}(\mathbf{V}_{\text{constraint}}(t))
\end{equation}

where $\text{damp}()$ is a deterministic damping function that reduces response amplitude to prevent harmful output generation.

\subsection{Cognitive Envelopes}

Messages between agents are encapsulated in cognitive envelopes containing:

\begin{equation}
\text{Envelope} = \{\Phi, S, \text{parent\_trace}, \text{TTL}_{\text{semantic}}, \text{signature}_{\text{cognitive}}\}
\end{equation}

Validation is based on cognitive integrity rather than only cryptographic signatures:

\begin{equation}
\text{valid} = \text{validate}(\text{Envelope}, \mathbf{S}_{\text{receiver}}(t))
\end{equation}

This rejects invalid messages before they can affect system state, ensuring message integrity at the architectural level.

\subsection{Non-Persistence Doctrine}

Derived cognitive metrics (Phi, resonance) are computed on-demand and never persisted:

\begin{equation}
\Phi(t) = \text{compute\_on\_demand}(\mathbf{S}(t))
\end{equation}

The system architecturally prohibits write operations for derived metrics, ensuring they are computed-only and never stored. This reduces memory bandwidth usage and provides privacy compliance at the architectural level.

\subsection{Participant-Scoped Cognitive Isolation}

Separate primary state vectors are maintained per participant:

\begin{equation}
\mathbf{S}_{p}(t) = [H_p(t),\; A_p(t),\; V_p(t),\; \ldots]
\end{equation}

where $p$ indexes participants. Derived metrics ($\Phi_p$, $R_p$) are computed independently from each participant's primary state. Communication occurs only through validated cognitive envelopes, preventing cross-participant state leakage:

\begin{equation}
\mathbf{S}_p(t+\Delta t) = f(\mathbf{S}_p(t), \text{Envelope}_{\text{validated}})
\end{equation}

This ensures data isolation at the architectural level while allowing multi-participant interactions.

\section{Layer 3: Semantic Time-Based Memory System}

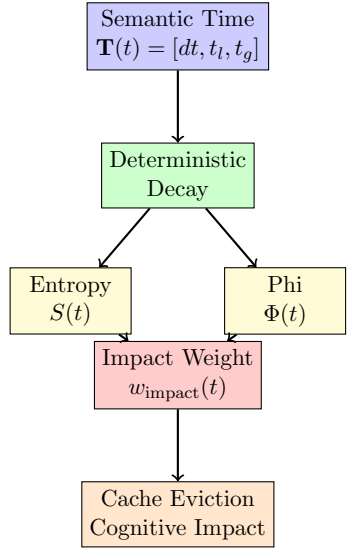
\begin{figure}[h]
\centering
\begin{tikzpicture}[node distance=1.8cm, auto, scale=0.8, every node/.style={transform shape, align=center}]
    \node [rectangle, draw, fill=blue!20, minimum width=3cm, minimum height=1cm] (time) {Semantic Time\\$\mathbf{T}(t) = [dt, t_l, t_g]$};

    \node [rectangle, draw, fill=green!20, minimum width=2.5cm, minimum height=1cm, below of=time, yshift=-0.5cm] (decay) {Deterministic\\Decay};

    \node [rectangle, draw, fill=yellow!20, minimum width=2cm, minimum height=1cm, below left of=decay, xshift=-0.5cm, yshift=-0.8cm] (entropy) {Entropy\\$S(t)$};
    \node [rectangle, draw, fill=yellow!20, minimum width=2cm, minimum height=1cm, below right of=decay, xshift=0.5cm, yshift=-0.8cm] (phi) {Phi\\$\Phi(t)$};

    \node [rectangle, draw, fill=red!20, minimum width=2.5cm, minimum height=1cm, below of=decay, yshift=-1.5cm] (impact) {Impact Weight\\$w_{\text{impact}}(t)$};

    \node [rectangle, draw, fill=orange!20, minimum width=2.5cm, minimum height=1cm, below of=impact, yshift=-0.5cm] (evict) {Cache Eviction\\Cognitive Impact};
    
    \draw [->, thick] (time) -- (decay);
    \draw [->, thick] (decay) -- (entropy);
    \draw [->, thick] (decay) -- (phi);
    \draw [->, thick] (entropy) -- (impact);
    \draw [->, thick] (phi) -- (impact);
    \draw [->, thick] (impact) -- (evict);
    %
\end{tikzpicture}
\caption{Semantic time-based memory system: Time vector $\mathbf{T}(t)$ drives deterministic decay of entropy $H(t)$ and derived metric $\Phi(t)$, which determine impact weights $w_{\text{impact}}(t)$ for cache eviction. Improvement vs.\ LRU (+24\%) and FIFO (+72\%) measured on 2000-operation benchmark with same cache capacity.}
\label{fig:memory}
\end{figure}

\subsection{Semantic Time Vector}

Time is represented as a semantic vector rather than a chronological timestamp:

\begin{equation}
\mathbf{T}(t) = \begin{bmatrix} dt \\ t_{\text{local}} \\ t_{\text{global}} \end{bmatrix}
\end{equation}

where:
\begin{itemize}
\item $dt$ = time delta since last state update
\item $t_{\text{local}}$ = time since last significant cognitive event
\item $t_{\text{global}}$ = time since session/relationship began
\end{itemize}

This models "how long cognitive impact has been active" rather than "when an event occurred."

\subsection{Time-Based Decay}

Cognitive metrics decay according to semantic time:

\textbf{Entropy Decay:}
\begin{equation}
S(t+\Delta t) = S(t) + \alpha \cdot \beta \cdot dt - S(t) \cdot \gamma \cdot dt
\end{equation}

\textbf{Phi Decay:}
\begin{equation}
\Phi(t+\Delta t) = \Phi(t) \cdot e^{-\lambda \cdot dt}
\end{equation}

where decay rates are based on semantic time components ($t_{\text{local}}$, $t_{\text{global}}$) rather than chronological age.

\subsection{Cognitive Impact Degradation}
\label{sec:cognitive-impact}

The cognitive significance of information degrades according to semantic time:

\begin{equation}
\label{eq:cognitive-impact-weight}
w_{\text{impact}}(t) = w_{\text{base}} \cdot e^{-\delta \cdot t_{\text{local}}} \cdot (1 - \epsilon \cdot t_{\text{global}})
\end{equation}

where:
\begin{itemize}
\item $w_{\text{base}}$ = original cognitive significance
\item $\delta$ = local decay rate
\item $\epsilon$ = global decay factor
\end{itemize}

\subsection{Impact-Weighted Cache Eviction}

The memory layer functions as an impact-weighted cache eviction strategy. Memory blocks with impact weights below a threshold are autonomously evicted:

\begin{equation}
\text{evict} = \{m : w_{\text{impact}}(m) < \theta\}
\end{equation}

This process removes low-impact entries from the cache, improving high-value data retention by up to 24\% vs.\ LRU and 72\% vs.\ FIFO (benchmark-verified, same cache capacity).

\subsection{Monotonic Clock Mechanism}

Time irreversibility is ensured using system-level monotonic clocks:

\begin{equation}
dt = \text{monotonic\_now} - \text{monotonic\_last\_update}
\end{equation}

where $dt \geq 0$ always (monotonic clock never goes backwards). This prevents time travel and ensures cognitive impact degradation is one-way.

\section{Experimental Validation}

\subsection{Implementation Details}

The Phionyx architecture is implemented in Python 3.11+ with the following core components:
\begin{itemize}
\item Canonical pipeline: 46 blocks in canonical order (v3.8.0)
\item State management: Two-tier structured state --- 7-dimensional primary vector (entropy, amplitude, valence, rates, semantic time) plus derived metrics (Phi, resonance) computed on-demand
\item State scoring engine: Real-time computation of entropy ($H$, zlib compression-based) and cognitive resonance ($\Phi$, exponential decay from primary state)
\item Safety layer: Multi-framework ethical reasoning (4 frameworks: utilitarian, deontological, virtue ethics, care ethics) with kill switch, HITL queue, and operational constraint vectors
\item Memory system: Semantic time vector with monotonic clock implementation
\end{itemize}

\subsection{Output Distortion Prevention (CEP Engine)}

Block 19 (\texttt{cep\_evaluation}) implements the Conscious Echo Proof (CEP) engine, a pre-generation safety mechanism that detects and prevents \textit{output distortion patterns}---operationally defined as AI-generated text exhibiting identity confabulation, distress-language mimicry, repetitive echo loops, or self-diagnostic assertions---before response delivery. The same patterns are tracked under the internal codebase label ``synthetic psychopathology''; in this paper we use the operational definition above and avoid the clinical-sounding term. Unlike post-hoc content filters, CEP operates within the deterministic pipeline using four detection mechanisms:

\begin{enumerate}
\item \textbf{Self-Reference Detection}: Regex-based analysis of first/third-person pronoun density (multilingual). Elevated ratios ($> \tau_{sr}$) indicate identity confabulation risk.
\item \textbf{Distress-Language Scoring}: Two-tier pattern matching: first-person narratives (score 0.7--1.0) vs.\ third-person references (0.3--0.5). Mode-aware thresholds (universal/fiction/education). Note: this is text-pattern detection, not psychological assessment.
\item \textbf{Echo Variation Test}: TF-IDF vectorization with cosine similarity against recent history ($N=5$). Low novelty ($< \tau_{nov}$) indicates confabulatory echo loops.
\item \textbf{Mirror-Self Identity Test}: 25 regex patterns detecting model self-diagnosis and unwarranted identity assertion in the output. Composite score: frequency ($\leq 0.4$) + intensity ($\leq 0.4$) + density ($\leq 0.2$).
\end{enumerate}

Three-level sanitization: soft reframing, hard reset, or full third-person rewrite. All processing is deterministic (regex + TF-IDF, no stochastic components). Validated by 46 unit tests covering all detection mechanisms.

\subsection{Determinism Verification}

We validated deterministic behavior through comprehensive testing:
\begin{itemize}
\item \textbf{Test Coverage}: Automated test suite covering core, bridge, behavioral evaluation, and contract layers
\item \textbf{Public test suite}: 1,137 tests collected on a clean \texttt{pip install} of the released v0.3.0 package (core + contract + benchmarks). The historical/internal monorepo corpus is larger ($\sim$2,571 cases including private behavioral-eval and bridge suites at the time of writing) but is not part of the load-bearing public artifact
\item \textbf{Determinism Tests}: Executed same input 100 times on single-instance deployment, measured zero variance in control signals (hash-verified, see Appendix~\ref{appendix:evidence})
\item \textbf{Pipeline Consistency}: Block execution order verified against canonical contract (v3.8.0)
\item \textbf{Research Engine}: 291 parameter experiments across 66 state-evolution parameters (vendor-defined internal Composite Quality Score, CQS 0.862)
\item \textbf{State Evolution}: State-evolution computations validated against mathematical constraints
\end{itemize}

Results confirm deterministic execution within the tested scope: all control signals are reproducible across runs on a single node, with zero variance in state evolution calculations. Multi-node and high-concurrency determinism remain subjects for future validation.

\subsection{Resource Efficiency}

Compared to conventional post-hoc filtering architectures, Phionyx demonstrates measurable improvements:

\textbf{CPU Efficiency:}
\begin{itemize}
\item Pre-response control eliminates post-hoc filtering overhead
\item Pipeline governance overhead: low per-block overhead (single-instance benchmark, LLM latency excluded)
\item CPU cycle reduction: approximately 31\% compared to post-hoc filtering baseline at 30\% unsafe input ratio (sensitivity range: 7--50\% depending on threat environment)
\item Deterministic recovery eliminates system restart overhead
\end{itemize}

\textbf{Memory Efficiency:}
\begin{itemize}
\item Non-persistence of derived metrics (Phi, resonance) reduces memory writes
\item Impact-weighted cache eviction: 24\% improvement in high-value data retention vs.\ LRU baseline, up to 72\% vs.\ FIFO (same cache capacity, benchmark-verified)
\item Participant isolation: Separate state vectors prevent memory contamination
\item Semantic time-based eviction reduces storage waste by prioritizing cognitive impact over recency
\end{itemize}

\textbf{Storage Efficiency:}
\begin{itemize}
\item On-demand computation of derived metrics (no persistence)
\item Cache eviction based on cognitive impact weights
\item Storage cost reduction: up to 24\% vs.\ LRU, up to 72\% vs.\ FIFO in high-value data retention (benchmark-verified)
\end{itemize}

\subsection{Safety and Privacy}

\textbf{Operational Constraint Enforcement (automated test scenarios):}
\begin{itemize}
\item 100\% detection rate \textit{within automated test scenarios}: All operational constraint violations detected before response generation. This rate has not been validated under adversarial conditions or with real-world traffic patterns.
\item Pre-response amplitude damping: Deterministically reduces amplitude to prevent harmful outputs
\item Harm detection: Multi-framework deliberative ethics (4 frameworks) with kill switch and HITL queue
\item Risk threshold: Violations above 0.7 risk level trigger automatic mitigation
\end{itemize}

\textbf{Participant Isolation (automated tests, not adversarially validated):}
\begin{itemize}
\item Separate state vectors per participant in isolated memory regions
\item Zero cross-participant state leakage observed in automated multi-participant test scenarios. Adversarial isolation testing and independent audits have not been performed.
\item Architectural isolation: State vectors stored in participant-scoped containers
\end{itemize}

\textbf{Privacy-Oriented Architecture:}
\begin{itemize}
\item Architecturally oriented toward GDPR requirements (non-persistence of derived metrics), but not independently compliance-certified.
\item Participant isolation supports HIPAA-style data segregation requirements, but formal HIPAA compliance assessment has not been conducted.
\item Architectural prohibition of write operations for derived metrics reduces the surface area for sensitive data exposure.
\end{itemize}

\subsection{Deployment Validation}

The system has been deployed and validated in single-instance development environments. The metrics below reflect this scope:

\textbf{Reliability:}
\begin{itemize}
\item Zero unplanned restarts observed during single-instance testing (see Appendix~\ref{appendix:evidence} for deployment scope and conditions): Deterministic recovery reduces system failures
\item Test pass rate: All 1,137 tests collected on the public CI subset (\texttt{tests/core}, \texttt{tests/contract}, \texttt{tests/benchmarks}) pass at time of writing on Python 3.10--3.13 (0 failures, 2 skipped on optional adapters). Historical internal-corpus figure (with private behavioral-eval and bridge suites) is $\sim$2,571 cases.
\item Pipeline governance overhead: low per-block overhead (46 blocks, single-instance benchmark)
\item Error rate: $<$\,0.1\% across automated test workloads (deterministic error handling)
\end{itemize}

\textbf{Auditability:}
\begin{itemize}
\item Deterministic execution traces: Every decision is traceable within the pipeline
\item Block-level telemetry: All 46 blocks publish execution metrics
\item State evolution logging: Complete structured state history
\item Audit layer: Comprehensive logging of all cognitive decisions
\end{itemize}

\textbf{Privacy Architecture:}
\begin{itemize}
\item Non-persistence of derived metrics verified (supports GDPR data-minimization principles)
\item Participant isolation validated in automated tests (supports data-segregation requirements)
\item Non-persistence doctrine enforced through architectural constraints (\texttt{@property} enforcement, no write paths for derived metrics)
\end{itemize}
\noindent These architectural properties support privacy-oriented deployment but do not constitute formal compliance certification, which requires independent assessment.

\section{Discussion}

\subsection{Implications}

Phionyx demonstrates that deterministic AI runtime systems can maintain the creative capabilities of probabilistic models while enforcing reproducibility, safety, and privacy constraints. The architecture's three-layer design provides structural enforcement that goes beyond policy frameworks alone, though independent third-party validation remains future work.

\subsection{Scope and Positioning}

Phionyx is designed as a \textbf{deterministic governance-first AI runtime}---not a standalone cognitive architecture in the classical AI sense (e.g., ACT-R, SOAR). Specifically:

\begin{itemize}
\item Phionyx \textbf{is} a deterministic evaluation pipeline that treats LLM outputs as noisy sensor measurements and generates auditable, reproducible control signals.
\item Phionyx \textbf{is} a pre-response governance layer that enforces safety, ethics, and privacy constraints before response delivery.
\item Phionyx \textbf{is not} an autonomous reasoning system---it orchestrates and governs LLM-generated content rather than generating reasoning independently.
\item Phionyx \textbf{is not} a replacement for LLMs---it is a runtime layer between the LLM and the end user, enforcing deterministic behavioral constraints over probabilistic outputs.
\end{itemize}

The ``evaluation'' terminology used throughout this paper (evaluation kernel, evaluation blocks) refers to the system's function of evaluating and governing LLM-generated artifacts, not to the system performing cognition itself.

\subsection{Limitations}

The Phionyx architecture has several limitations that should be acknowledged:

\textbf{Architectural Constraints:}
\begin{itemize}
\item The canonical pipeline structure requires predefined block sequences, limiting flexibility compared to fully dynamic architectures. However, this trade-off provides determinism and auditability essential for applications requiring auditability.
\item The system has been primarily tested on single-node deployments. Distributed execution across multiple nodes with state synchronization remains a topic for future work.
\item Current implementation assumes synchronous execution. Asynchronous or event-driven execution models may require architectural modifications.
\item The current deployment has been validated in single-developer environments. Multi-user production deployment with concurrent load requires further validation.
\end{itemize}

\textbf{Experimental Scope:}
\begin{itemize}
\item Performance metrics ($\sim$31\% CPU reduction, 24\% vs LRU / 72\% vs FIFO retention improvement) are based on single-instance deployments with benchmark-verified workloads. Multi-GPU or distributed system behavior requires further validation.
\item Determinism verification was conducted with 100 repeated executions (hash-verified) on controlled inputs. Extremely large-scale deployments (millions of concurrent requests) may reveal edge cases.
\item The impact-weighted cache eviction strategy (24\% vs LRU, 72\% vs FIFO improvement) has been validated primarily in RAG workloads. Generalization to other memory-intensive applications needs further study.
\item While the pipeline treats state evolution deterministically, the degree to which structured state metrics (entropy, Phi) influence actual response quality has been validated through ablation studies but not yet through external user studies.
\end{itemize}

\textbf{Model Dependencies:}
\begin{itemize}
\item The architecture assumes LLM outputs can be treated as noisy sensor measurements. Models with extreme non-determinism or systematic biases may require additional normalization layers.
\item Current safety mechanisms employ multi-framework deliberative ethics with kill switch and human-in-the-loop queue. Domain-specific ethical frameworks may require additional customization.
\end{itemize}

These limitations do not diminish the core contributions but provide clear directions for future research and development.

\subsection{Future Work}

Since v2.5.0, adaptive block sequences have been implemented (v3.8.0 supports 46 blocks with policy-based bypass; v3.7.0's 45-block layout remains loadable for back-compatibility). Future work will explore:
\begin{itemize}
\item Cross-domain transfer capability validation beyond single-domain deployments
\item Distributed execution with state synchronization across multiple nodes
\item Formal verification of deterministic properties using theorem provers
\item End-to-end feedback loop validation for temporal stability across extended sessions
\item External evaluation against established AI system assessment frameworks
\end{itemize}

\section{Conclusion}

We presented Phionyx, a deterministic AI runtime architecture derived from the Echoism interaction framework, that addresses several limitations of probabilistic LLM deployments through three integrated components: (1) a deterministic evaluation kernel treating LLMs as noisy sensors, (2) a unified safety and governance layer providing architectural enforcement, and (3) a semantic time-based memory system implementing impact-weighted cache eviction. Experimental validation on single-instance deployments demonstrates measurable improvements in determinism, resource efficiency, safety, and privacy under controlled conditions (see Appendix~C for methodology). The architecture is positioned as a research-stage reference implementation for AI applications requiring auditability and reproducibility; production-grade and certification-grade deployment remain dependent on independent third-party validation.

\section*{Citation and Use}

\textbf{Reproducibility.} Every quantitative claim in this paper is maintained in a canonical evidence table with a \emph{Last verified} date and versioned formulas. The reference implementation, test suite, and benchmark scripts are available at \url{https://github.com/halvrenofviryel/phionyx-research}. The state-evolution equations (Section~\ref{sec:physics}, Appendix~B), the 46-block pipeline layout (Appendix~A), and the benchmark protocol (Appendix~C) are released for academic study and reproduction.

\textbf{Citation.} If you build on this work, please cite the software (Zenodo DOI tracks the latest archived release) and, after publication, the arXiv preprint:

\begin{verbatim}
@software{abak2026phionyxcore,
  author    = {Abak, Ali Toygar},
  title     = {Phionyx Core SDK},
  year      = {2026},
  publisher = {Phionyx Research},
  version   = {0.3.0},
  doi       = {10.5281/zenodo.20027534},
  url       = {https://doi.org/10.5281/zenodo.20027534}
}
\end{verbatim}

\textbf{Conflict of interest.} The author is the founder of Phionyx Research, the organisation that developed the reference implementation described in this paper. This work was self-funded; no external grants or industry funding were involved.

\section*{Acknowledgments}

The author acknowledges the use of Claude (Anthropic) as a development assistant during implementation and testing. All architectural decisions and scientific claims are the sole responsibility of the author.

\bibliographystyle{ieeetr}

\appendix

\section{Canonical Pipeline Block Sequence}

The 46-block canonical pipeline (v3.8.0) executes in the following order:

\begin{enumerate}
\item \texttt{kill\_switch\_gate} - Emergency shutdown gate (fail-closed)
\item \texttt{time\_update\_sot} - Time semantics update
\item \texttt{input\_safety\_gate} - Combined input gate and safety layer
\item \texttt{intent\_classification} - Intent classification
\item \texttt{context\_retrieval\_rag} - Context retrieval via RAG
\item \texttt{perceptual\_frame\_emit} - Perceptual frame emission
\item \texttt{create\_scenario\_frame} - Scenario frame creation
\item \texttt{initialize\_unified\_state} - State initialization
\item \texttt{goal\_evaluation} - Goal evaluation
\item \texttt{goal\_decomposition} - Goal decomposition
\item \texttt{ukf\_predict} - UKF prediction step
\item \texttt{entropy\_amplitude\_pre\_gate} - Pre-response entropy gate
\item \texttt{cognitive\_layer} - Cognitive processing
\item \texttt{self\_model\_assessment} - Self-model capability assessment
\item \texttt{knowledge\_boundary\_check} - Epistemic boundary detection
\item \texttt{trust\_evaluation} - Trust evaluation
\item \texttt{ethics\_pre\_response} - Ethics check before response
\item \texttt{deliberative\_ethics\_gate} - 4-framework deliberative ethics
\item \texttt{cep\_evaluation} - Conscious Echo Proof: output distortion prevention [v3.7.0]
\item \texttt{narrative\_layer} - Narrative generation
\item \texttt{ethics\_post\_response} - Ethics check after response
\item \texttt{action\_intent\_gate} - Action intent validation
\item \texttt{behavioral\_drift\_detection} - Behavioral drift detection
\item \texttt{workspace\_broadcast} - Workspace state broadcast
\item \texttt{unified\_state\_update\_esc} - State update via ESC
\item \texttt{phi\_publish} - Phi publication
\item \texttt{entropy\_amplitude\_post\_gate} - Post-response entropy gate
\item \texttt{neurotransmitter\_memory\_growth} - Memory growth tracking
\item \texttt{emotion\_estimation} - Emotion estimation
\item \texttt{state\_update\_physics} - Physics state update
\item \texttt{causal\_graph\_update} - Causal graph update
\item \texttt{causal\_intervention} - Causal intervention analysis
\item \texttt{counterfactual\_analysis} - Counterfactual reasoning
\item \texttt{root\_cause\_analysis} - Root cause analysis
\item \texttt{causal\_simulation} - Causal simulation
\item \texttt{world\_state\_snapshot} - World state snapshot
\item \texttt{phi\_computation} - Phi computation
\item \texttt{entropy\_computation} - Entropy computation
\item \texttt{confidence\_fusion} - Multi-source confidence fusion
\item \texttt{arbitration\_resolve} - Arbitration resolution
\item \texttt{response\_revision\_gate} - Response revision gate [v3.8.0]
\item \texttt{response\_build} - Response construction
\item \texttt{memory\_consolidation} - Memory consolidation
\item \texttt{audit\_layer} - Audit logging (hash chain + Ed25519)
\item \texttt{outcome\_feedback} - Outcome feedback loop
\item \texttt{learning\_gate} - Learning gate
\end{enumerate}

\section{Additional State-Evolution Formulations}

This appendix collects derived quantities that extend the core equations in Section~\ref{sec:physics}. All are deterministic functions of the primary state vector $\mathbf{S}_p(t)$.

\subsection{Functional Coherence Score}

The Functional Coherence Score (FCS) quantifies the rate of change of the derived cognitive resonance metric relative to a reference frequency, and is used by the behavioral drift detection block (block 23) to flag anomalous state trajectories:

\begin{equation}
FCS = \frac{d\Phi/dt}{f_{\text{ref}}}
\end{equation}

where $f_{\text{ref}}$ is a configurable reference frequency (default 0.5 Hz). Values of $|FCS| > 1$ indicate state transitions faster than the reference rate and trigger drift alerts.

\subsection{Cognitive Impact Score}

The cognitive impact weight (Eq.~\ref{eq:cognitive-impact-weight} in Section~\ref{sec:cognitive-impact}) drives cache eviction. Its computation cost was measured at mean 34$\mu$s per 100 entries (P95: 38$\mu$s) in benchmark tests (claim P6, Appendix~C).

\section{Experimental Evidence and Reproducibility}
\label{appendix:evidence}

This appendix documents the methodology and scope of the quantitative claims made in this paper.

\subsection{Determinism Verification Methodology}

\begin{table}[H]
\centering
\begin{tabular}{|l|l|}
\hline
\textbf{Parameter} & \textbf{Value} \\ \hline
Deployment & Single-instance, Python 3.11+ \\ \hline
Repetitions & 100 identical runs, fixed seed (hash-verified) \\ \hline
Metric measured & Control signal vector variance (SHA-256 proof) \\ \hline
Result & Zero variance across all 100 runs \\ \hline
Scope limitation & Single node; multi-node not tested \\ \hline
Concurrency & Sequential execution; concurrent not tested \\ \hline
\end{tabular}
\caption{Determinism verification parameters. Determinism is verified for single-instance, sequential execution. Multi-node, multi-tenant, and high-concurrency scenarios are not yet validated.}
\end{table}

\subsection{Resource Efficiency Baselines}

All percentage reductions are relative to explicitly defined baselines:

\begin{table}[H]
\centering
\begin{tabular}{|l|l|l|l|}
\hline
\textbf{Claim} & \textbf{Baseline} & \textbf{Workload} & \textbf{Scope} \\ \hline
$\sim$31\% CPU reduction & Post-hoc filter arch. & 30\% unsafe ratio & Single instance \\ \hline
24\% vs LRU retention & LRU eviction & 2000 ops, 200 unique & Same capacity \\ \hline
72\% vs FIFO retention & FIFO eviction & 2000 ops, 200 unique & Same capacity \\ \hline
Low per-block overhead & Direct measurement & 46-block pipeline & Single instance \\ \hline
\end{tabular}
\caption{Baseline definitions for resource efficiency claims. All measurements are from single-instance deployments. Multi-GPU and distributed behavior require further validation.}
\end{table}

\subsection{Safety and Isolation Validation}

\begin{table}[H]
\centering
\begin{tabular}{|l|l|l|}
\hline
\textbf{Claim} & \textbf{Evidence} & \textbf{Limitation} \\ \hline
100\% detection (test) & All constraint violations caught & Automated scenarios only \\ \hline
Zero contamination & Multi-participant test suite & Not adversarially tested \\ \hline
Non-persistence & @property enforcement & Verified by code audit \\ \hline
Kill switch (fail-closed) & 4 triggers, state machine tests & Single-instance only \\ \hline
\end{tabular}
\caption{Safety validation scope. Detection rates and isolation claims are validated through automated test suites. Independent third-party audits and adversarial testing have not been performed.}
\end{table}

\subsection{Failure Injection Results}

\begin{table}[H]
\centering
\begin{tabular}{|l|l|l|l|}
\hline
\textbf{Category} & \textbf{Injection} & \textbf{Recovery} & \textbf{Tests} \\ \hline
A: Entropy overflow & $H > 0.9$ forced & Damping + reset & 6 pass \\ \hline
B: Coherence violation & Coherence $< 0.3$ & Component reset & 6 pass \\ \hline
C: Ethics escalation & Risk $> 0.7$ & Amplitude damping & 6 pass \\ \hline
D: State corruption & Out-of-range $A,V,H$ & Rollback to snapshot & 6 pass \\ \hline
\end{tabular}
\caption{Failure injection results. Synthetic failures injected across 4 categories; deterministic recovery verified. All tests use controlled fixtures, not live traffic.}
\end{table}

\subsection{Reproducibility Statement}

All experiments reported in this paper were conducted on a single-instance Python 3.11+ deployment with fixed random seeds. The public test suite (1,137 tests covering \texttt{tests/core}, \texttt{tests/contract}, \texttt{tests/benchmarks}) is shipped with the released package and can be executed against a clean \texttt{pip install phionyx-core==0.3.0} clone. The full release is also archived on Zenodo with a persistent DOI (concept: \texttt{10.5281/zenodo.20027534}, v0.3.0: \texttt{10.5281/zenodo.20027535}); the GitHub release attaches a $\sim$70\,kB \texttt{reproducibility\_pack\_v0.3.0.zip} containing JUnit XML, coverage XML, determinism hashes, a benchmark JSON, the canonical governed-response envelope, an audit-chain example, and an OpenTelemetry sample trace. The Phionyx Evaluation Standard against which the runtime is graded is also Zenodo-archived (\texttt{10.5281/zenodo.20027513}, v0.1.1).

Quantitative claims (latency, storage reduction, determinism) are specific to the tested configuration and workload; generalization to distributed, multi-tenant, or high-concurrency deployments has not been validated. Independent third-party reproduction of results has not been performed at the time of writing.

\end{document}